\documentclass[11pt]{article}

\usepackage[preprint]{acl}

\usepackage{times}
\usepackage{latexsym}
\usepackage{xcolor}
\usepackage{xspace}
\usepackage{amsmath}
\usepackage{amssymb}
\usepackage{bbm}
\usepackage{booktabs}
\usepackage{threeparttable}
\usepackage{colortbl}
\usepackage{graphicx}
\usepackage{multirow}
\usepackage{soul}
\usepackage{breqn}
\usepackage{array}
\usepackage{tabularx}
\usepackage{xurl}
\usepackage{hyperref}

\usepackage[T1]{fontenc}

\usepackage[utf8]{inputenc}

\usepackage{microtype}

\usepackage{inconsolata}

\usepackage{graphicx}

\newcommand{\mumoinstruct}{\texttt{MuMOInstruct}\xspace}

\newcommand{\gellmo}{\mbox{GeLLM\textsuperscript{3}O}\xspace}

\newcommand{\gellmomfg}{\mbox{GeLLM\textsuperscript{3}O-MFG}\xspace}

\newcommand{\gellmoBaseRL}{\mbox{GeLLM\textsuperscript{3}O-RL}\xspace}

\newcommand{\gellmoPSix}{\mbox{GeLLM\textsuperscript{3}O-P(6)\textsubscript{\textrm{Llama}}}\xspace}

\newcommand{\framework}{{MFG}\xspace}

\newcommand{\MOMO}{\mbox{$\mathop{\text{MOMO}}\limits$}\xspace}
\newcommand{\policy}{\ensuremath{\pi_\theta}\xspace}

\newcommand{\turn}{\mbox{$\mathop{\tau}\limits$}\xspace}

\newcommand{\feedV}{\mbox{$\mathop{\mathbf{f}}\limits$}\xspace}

\newcommand{\source}{\mbox{$\mathop{C_0}\limits$}\xspace}

\newcommand{\property}{\mbox{$\mathop{p}\limits$}\xspace}
\newcommand{\propertyset}{\mbox{$\mathop{\mathcal{P}}\limits$}\xspace}

\newcommand{\sr}{\texttt{SR}\xspace}

\newcommand{\srZeroFive}{\texttt{SR}_{\scalebox{0.7}{\texttt{0.5}}}\xspace}
\newcommand{\srZeroSix}{\texttt{SR}_{\scalebox{0.7}{\texttt{0.6}}}\xspace}

\newcommand{\similarity}{\texttt{Sim}\xspace}
\newcommand{\RI}{\texttt{RI}\xspace}

\newcommand{\qwen}{\mbox{Qwen3.5-9B}\xspace}
\newcommand{\ministral}{\mbox{Ministral-3-8B}\xspace}

\newcommand{\biomedgpt}{\mbox{BioMedGPT-Mol}\xspace}
\newcommand{\llasmol}{\mbox{LlaSMol}\xspace}

\title{MFG: Multi-turn Feedback Guided Molecular Optimization}

\author{
 \textbf{Trieu Nguyen\textsuperscript{1}},
 \textbf{Frazier N. Baker\textsuperscript{1}},
 \textbf{Ruoxi Gao\textsuperscript{1}},
 \textbf{Daniel Adu-Ampratwum\textsuperscript{4}},
 \textbf{Xia Ning\textsuperscript{1,2,3,4}}
\\
\\
 \textsuperscript{1}Department of Computer Science and Engineering, The Ohio State University
 \\
 \textsuperscript{2}Department of Biomedical Informatics, The Ohio State University
 \\
 \textsuperscript{3}Translational Data Analytics Institute, The Ohio State University
 \\
 \textsuperscript{4}Division of Medicinal Chemistry and Pharmacognosy, The Ohio State University
\\
 \small{
   \textbf{Correspondence:} \href{mailto:ning.104@osu.edu}{ning.104@osu.edu}
 }
}
\ttfamily
\begin{document}
\maketitle
\begin{abstract}
Multi-objective molecular optimization is a critical process in drug discovery that aims to modify a given molecule to simultaneously improve multiple properties while maintaining a similar overall structure. 
This process often requires iterative modifications guided by computational or experimental feedback.
Recent large language model (LLM)-based methods have begun to explore this iterative approach; however, these methods are not explicitly trained to leverage such feedback, while existing approaches primarily focus on single-turn molecular optimization.
To address this gap, we introduce Multi-turn Feedback Guided (MFG) molecular optimization, a reinforcement learning framework that trains an LLM to progressively optimize molecules using feedback from previous turns. 
MFG introduces a feedback vector capturing both property improvements and source molecule similarity, alongside a feedback-aligned reward function that promotes progressive improvement across turns.
We apply MFG over $\text{GeLLM}^{3}\text{O}$, a state-of-the-art LLM model for multi-objective molecular optimization, evaluate $\text{GeLLM}^{3}\text{O-MFG}$ on ten multi-objective molecular optimization tasks. 
The results demonstrate that $\text{GeLLM}^{3}\text{O-MFG}$ achieves the highest average similarity-constrained success rate (81.50\%), outperforming the base model $\text{GeLLM}^{3}\text{O}$ ($70.81\%$).

\end{abstract}

\section{Introduction}
\label{sec:introduction}
Multi-objective molecular optimization (MOMO) \cite{molecule_optimization} is a critical process in drug discovery \cite{principles_dd} that aims to modify a given molecule to simultaneously improve multiple properties while maintaining a similar overall structure.
Such improvements are necessary because even promising molecules may suffer from poor solubility, insufficient selectivity, or toxicity that hinders their development into viable drugs.
Achieving these improvements often requires iterative structural modifications to balance multiple, potentially competing molecular properties.
By iteratively reasoning on how to modify molecules,
chemists can rely on not only their expert knowledge, but also computational \cite{multi_objective_opt} or experimental feedback.
Large language models (LLMs) have shown promise on similar iterative tasks through multi-turn reasoning~\cite{naveed2025comprehensive},
providing a promising mechanism to mimic this approach to MOMO.

Recent efforts \cite{ye2025drugassist, zuo2025biomedgpt} have used LLMs for MOMO.
Several methods \cite{liu2023chatgpt, le2024agentdrug} incorporated iterative feedback at inference-time,
prompting LLMs to modify molecules over multiple turns while incorporating feedback on the modified molecules of previous turns.
However, these methods are not trained on how to leverage such feedback.
Instead, they rely only on the LLMs' general-purpose pretraining and basic reasoning capabilities.
Other methods have used supervised finetuning methods \cite{dey-etal-2025-gellm3o} and reinforcement learning (RL) \cite{liu2025drugimprovergpt} to train models for MOMO.
However, training is limited to non-iterative, single-turn generation,
with no specialized training on multi-turn feedback.

Here, we introduce \framework, a \textbf{M}ulti-turn \textbf{F}eedback \textbf{G}uided framework for MOMO.
\framework formulates MOMO as a multi-turn problem with feedback, using RL to train an LLM to iteratively optimize molecules by leveraging feedback from previous turns.
\framework introduces a feedback vector that captures both property improvements and similarity to the given source molecule at each turn,
providing dynamically updated feedback to guide the LLM policy throughout the optimization process.
\framework's RL training is guided by a reward function that aligns with this feedback while also encouraging progressive improvement across turns.
During inference, \framework can use its trained model to generate multiple candidate molecules at each turn and select the best molecule for further turns,
carefully balancing property improvement and similarity.
Together, \framework serves as a powerful new framework for MOMO with feedback. 

%
%


To evaluate \framework, we instantiate \gellmomfg, using \gellmo, a state-of-the-art molecular optimization LLM, as the base model for \framework's RL training.
We compare \gellmomfg against \gellmo and other existing molecular optimization baselines
on the \mumoinstruct dataset~\cite{dey-etal-2025-gellm3o}, a benchmark featuring diverse property combinations and both in-distribution and out-of-distribution MOMO tasks.
%
Across all tasks,
\gellmomfg achieves the highest 
average similarity-constrained success rate (81.50\%), outperforming the best baseline (\biomedgpt, $77.81\%$) and \gellmo ($70.81\%$).
%
%
An ablation study shows that multi-turn feedback and RL training each contribute a distinct advantage to \framework.
Additionally, a case study illustrates how \gellmomfg adapts its optimization strategy based on feedback.
\gellmomfg is publicly available at \url{https://anonymous.4open.science/r/MFG-85BD/}.

\section{Related Work}
\label{sec:related_work}

\paragraph{RL for Molecular optimization}
Existing molecular optimization LLMs such as \gellmo~\cite{dey-etal-2025-gellm3o} have explored instruction tuning of general-purpose LLMs on large-scale \MOMO tasks (see Appendix~\ref{app:related_work} for a summary of more related work).
However, this supervised learning approach may limit model's ability to explore chemical space beyond the distribution of training examples.
Recent works have leveraged RL to further align molecular optimization LLMs with desired molecule properties.
For example, DrugImproverGPT \cite{liu2025drugimprovergpt} employed reward-based RL to explore molecule modifications for improving multiple molecule properties.
However, the sparsity of high-reward molecules in chemical space can make effective exploration challenging. 
To address this challenge, RePO \cite{li2026reference} leveraged reference optimized molecules to provide LLMs with templates of successful optimization.
Despite different designs of RL methods, existing works primarily optimize source molecules in a single turn, treating each optimization as an independent process.
Our work considers a new setting in which molecules are iteratively optimized over multiple turns, and develops a novel multi-turn training framework for this setting.
To the best of our knowledge, our work is the first to improve molecular optimization LLMs through multi-turn RL.

\section{\framework Framework}
\label{sec:framework}

\begin{figure*}[t]
    \centering
    \includegraphics[width=\textwidth]{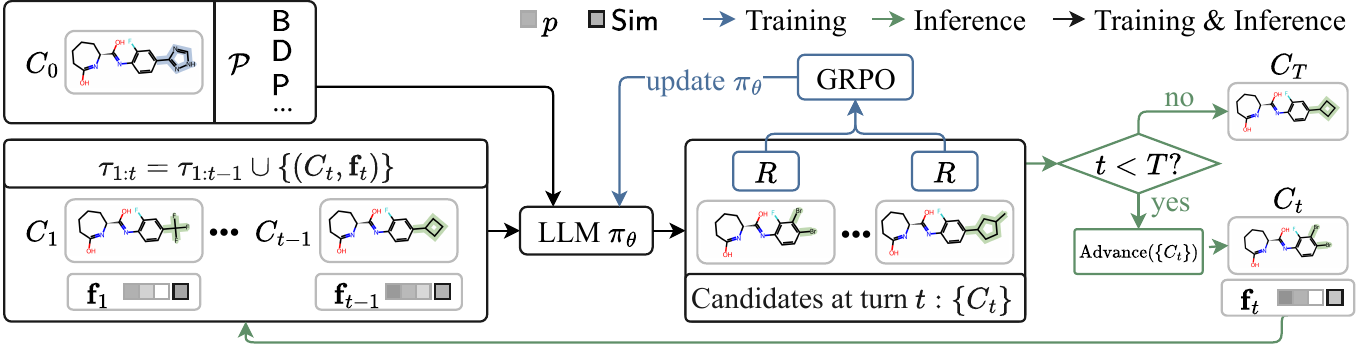}
    \caption{Overall figure of Multi-turn Feedback Guided (\framework) \MOMO.}
    \label{fig:overview}
\end{figure*}

\subsection{Problem Formulation}
\label{sec:framework:problem_formulation}
Multi-objective molecular optimization (\MOMO) seeks to modify a source molecule \source into a structurally similar molecule C while improving $m$ properties $\propertyset=(\property_1, \property_2, \cdots, \property_m)$.
Existing LLM-based approaches commonly
formulate this task as
a single-turn problem,
where an LLM directly generates $C$ in a single response when prompted with \source and \propertyset.
However, this single-turn approach offers limited opportunity to adjust the optimization strategy.
MOMO can be very challenging, as structural changes that improve one property may degrade another, and improvement patterns may not remain consistent across different \source.

To address this challenge, we reformulate \MOMO as a multi-turn problem with feedback, where an LLM iteratively generates optimized molecules for a given \source and \propertyset, adjusting its strategies based on previous turn's property change and similarity to \source.
%
%
Under this formulation, we consider the LLM with parameters $\theta$ as a policy \policy that decides how to generate a molecule $C_{t}$ at any
turn $t$ given \source, \propertyset, and the sequence of previous turns up to $t-1$, denoted as $\turn_{1:t-1}$.
Each turn $\turn_{t}$ is defined as $\turn_{t} = (C_t, \feedV_t)$, where $\feedV_t \in \mathbb{R}^{m+1}$ is a feedback vector measuring the quality of $C_t$, with the first $m$ entries indicating degradation or improvement in each property objective $\property \in \propertyset$,
and the final entry representing the similarity to $\source$.
This provides \policy with valuable insight into each specific \MOMO problem, allowing it to adjust its molecular optimization strategy through trial-and-error.
An optimal \policy leverages this feedback across all turns to maximize the average improvement on the property set $\propertyset$ and maintain strong similarity to \source.

To obtain such a policy, we formulate the following optimization problem:
\begin{equation}
\label{eq:objective}
    \max_{\policy}
    \mathbb{E}_{t, {\scriptsize C_t \sim \policy(\cdot|\source, \propertyset, \turn_{1:t-1})}}
    \left[
    R(C_t)
    \right],
\end{equation}
where $t \in \{1, \ldots, T\}$ is the number of turns up to maximum $T$ and $R(C_t)$ is a reward function specifically designed for \MOMO (defined in Section~\ref{sec:framework:reward_func}).
This formulation distributes the optimization of multiple property objectives across turns, with reward signals guiding progressive improvement throughout the process.
An overview of \framework is illustrated in Figure~\ref{fig:overview}.

\subsection{\framework Training}
\label{sec:framework:training}
\framework trains \policy to leverage feedback for multi-turn \MOMO by maximizing an expected reward $R(C_t)$ (Section~\ref{sec:framework:reward_func}).
Specifically, we use group-relative policy optimization (GRPO) \cite{shao2024deepseekmath} to train $\pi_\theta$ by sampling multiple candidate molecules $C_t$ at each turn $t$ conditioned on $C_0$, $\mathcal{P}$, and $\tau_{1:t-1}$, and updating $\pi_\theta$ based on their relative rewards. 
To optimize the expectation over $t$ in Eq. 1, MFG samples $t \sim \mathrm{Uniform}(1,T)$ during training, ensuring that $\pi_\theta$ is trained to leverage feedback at every turn in multi-turn MOMO. 
\framework employs on-policy training, where the previous turns $\tau_{1:t-1}$ are generated using the same $\pi_\theta$ that generates $C_t$. 
However, \framework treats $\tau_{1:t-1}$ as fixed conditioning context rather than optimizing its generation based on $R(C_t)$. 
Together, these design choices allow $\pi_\theta$ to learn effective feedback-guided molecular optimization strategies by comparing candidate molecules generated under the same conditioning context.
As $\pi_\theta$ is updated during training, the conditioning of previous turns $\tau_{1:t-1}$, which are generated on-policy and used as conditioning context, also reflect the policy's changing output distribution.

\subsubsection{Candidate Advancement Criterion}
\label{sec:framework:priority}
To encourage progressive improvements across turns, \framework defines a candidate selection criterion $\operatorname{Advance}(\cdot)$ with two purposes: (1) during training, identifying the best molecule from previous turns as a reference for progressive improvement; and (2) during inference, selecting a promising molecule to carry forward, along with its feedback, to guide subsequent optimization.
Specifically, $\operatorname{Advance}(\cdot)$ is a hierarchical selection criterion that balances improvements across multiple target properties with structural similarity to $\source$ through prioritized comparisons.
It prioritizes molecules satisfying a predefined similarity threshold $\delta$ to $\source$, followed by molecules that improve all target properties. 
When no candidate improves all properties, those improving more properties are preferred, preventing large gains in one property from outweighing broader improvements. 
Together, $\operatorname{Advance}(\cdot)$ prioritizes the breadth of property improvements over their magnitude while preserving structural similarity (see Appendix~\ref{app:selection} for details).


\subsubsection{Reward Function}
\label{sec:framework:reward_func}
\framework defines a reward function $\mbox{$R(C_t \mid C_0, \propertyset, C_{t-1}^*)$}$ that encourages property improvement, structural similarity to $C_0$, and progressive improvements over the best molecule from the previous turns.
The goal is to prioritize structural similarity as a prerequisite for rewarding property improvements, while encouraging the model to progressively generate better molecules across turns.
Specifically, $R(C_t)$ (with conditioning omitted for simplicity) is defined as:
\begin{equation}
\label{reward_function}
R(C_t) =
\begin{cases}
s - 2,
& \text{if } s \leq \delta, \\
R_{\mathrm{base}}(C_t) + R_{\mathrm{best}}(C_t),
& \text{otherwise},
\end{cases}
\end{equation}
where $s = \operatorname{Sim}(C_t, C_0)$ denotes the Tanimoto \cite{bajusz2015tanimoto} similarity between $C_t$ and $C_0$.

As shown in Eq~\ref{reward_function}, when $s \leq \delta$, $R(C_t)$ assigns a similarity-dependent penalty, without accounting for property improvements .
This prevents large property gains from compensating for excessive structural deviations.
Moreover, this penalty complements our algorithm choice, GRPO, which updates the policy based on relative rewards among candidates in a group. 
Even when all candidates within a group breach the similarity threshold, those with higher similarity to $\source$ receive higher rewards, providing a learning signal that encourages the model to generate more structurally similar molecules.

For molecules that meet the similarity threshold, we introduce $R_{\mathrm{base}}(\cdot)$ to incentivize both property improvements and similarity to $C_0$, and is defined as follows:
\begin{equation}
R_{\mathrm{base}}(C_t)
= \lambda(C_t) \left(\mathcal{G}(C_t) + \mathcal{S}(C_t)\right).
\label{eq:reward_definition}
\end{equation} 
Here, $\mathcal{G}(\cdot)$ and $\mathcal{S}(\cdot)$ measure the average property improvement on \propertyset and the similarity to \source, respectively, each normalized by the corresponding value of the reference optimized molecule (details in Section~\ref{sec:experimental_setup:dataset}).
%
%
The normalization addresses differences in the numerical scales of different properties and encourages \policy to generate molecules outperforming the reference molecule.
$\lambda(C_t) \in [1,2)$ amplifies the reward only when $C_t$ improves all target properties over $C_0$, with a larger bonus for higher structural similarity. 
Unlike $\mathcal{G}(\cdot)$, which measures average property improvement, $\lambda(\cdot)$ encourages simultaneous improvement across competing properties rather than large gains in a few at the expense of others.
Details of $\mathcal{G}(\cdot)$, $\mathcal{S}(\cdot)$, and $\lambda(\cdot)$ are provided in Appendix~\ref{app:reward_function}.

In addition to evaluating $C_t$ with respect to $C_0$, $R_{\mathrm{best}}(\cdot)$ in Eq~\ref{reward_function} compares $C_t$ with the best molecule $C^*_{t-1}$ from the previous $t-1$ turns, identified using $\operatorname{Advance}(\cdot)$ described in Section~\ref{sec:framework:priority}.
%
%
$R_{\mathrm{best}}(\cdot)$ is defined as follows:
\begin{equation}
R_{\mathrm{best}}(C_t)
=
\left[
R_{\mathrm{base}}(C_t)
-
R_{\mathrm{base}}(C_{t-1}^*)
\right]_+,
\label{eq:reward_best}
\end{equation}
where $[\cdot]_+ = \max(\cdot, 0)$ denotes the
positive-part operator.
This sub-reward provides a bonus when $C_t$ achieves a higher base reward than $C_{t-1}^*$, encouraging progressive improvement across turns.


\subsection{\framework Inference}
\label{sec:framework:inference}
At inference time, \framework performs iterative molecular optimization using the trained policy \policy, conditioned on the source molecule \source, target property set \propertyset, initially empty set of previous turns $\tau$.
At each turn $t$, \framework samples $N$ candidates and applies the selection criterion $\operatorname{Advance}(\cdot)$ introduced in Section~\ref{sec:framework:priority} to obtain $C_t=\operatorname{Advance}(\{C_t^1,\ldots,C_t^N\})$.
The selected molecule and its feedback vector $\mathbf{f}_t$, capturing property changes and structural similarity relative to \source, form the updated $\tau_{1:t}=\tau_{1:t-1}\cup\{(C_t,\mathbf{f}_t)\}$.
This process repeats until turn $T$, with each selected molecule and its feedback incorporated into the conditioning context for the next turn.
The final optimized molecule $C_T$ is then selected from all $T \times N$ generated candidates according to the criteria described in Section~\ref{sec:experimental_setup:evaluation_metrics}.


Unlike the final output selection, $\operatorname{Advance}(\cdot)$ contributes to feedback construction during inference by determining which intermediate molecule and its associated feedback are carried forward to subsequent turns.
This selection aims to carry forward a promising intermediate molecule that provides a meaningful direction for subsequent optimization, allowing future turns to build on partial progress.
Thus, $\operatorname{Advance}(\cdot)$ supports feedback-guided MOMO by limiting the propagation of unfavorable candidates and retaining informative feedback that helps the model adapt across turns.

\section{Experimental Settings}
\label{sec:experimental_setup}

\subsection{\gellmomfg}
\label{sec:experimental_setup:training_settings}

We use \framework to train \gellmomfg, a multi-turn \MOMO model based on \gellmoPSix, a state-of-the-art single-turn \MOMO model.
\gellmoPSix is particularly suitable for training with \framework
because its previous training on single-turn \MOMO has allowed it to acquire substantial knowledge of structure-property relationships.
%
%
%
%
Implementation details of \gellmomfg are provided in Appendix~\ref{app:implementation}.
 
%
%

\subsection{Dataset}
\label{sec:experimental_setup:dataset}
\gellmomfg is trained and evaluated using the \mumoinstruct benchmark dataset following the training/testing split established by \gellmo. 
\mumoinstruct includes multi-property optimization tasks on combinations of six target properties: blood-brain barrier permeability (\texttt{B}), dopamine receptor D2 binding affinity (\texttt{D}), human intestinal absorption (\texttt{H}), mutagenicity (\texttt{M}), penalized logP, a measure of lipophilicity (\texttt{P}), and quantitative estimate of drug-likeness (\texttt{Q}). 
The optimization objective is to increase the values of \texttt{B}, \texttt{D}, \texttt{H}, \texttt{P}, and \texttt{Q}, while decreasing that of \texttt{M}.
Each task consists of a set of source molecules, paired with their corresponding reference optimized molecules that we use for reward normalization in Eq~\ref{eq:reward_definition}. 
Specifically, the property improvements achieved by the reference optimized molecule relative to the source molecule are used to normalize $\mathcal{G}(\cdot)$, while its structural similarity to the source molecule is used to normalize $\mathcal{S}(\cdot)$ (details in Appendix~\ref{app:reward_function}).
%
%
%


The evaluation set consists of five in-distribution (IND) tasks (\texttt{BDP}, \texttt{BDQ}, \texttt{BPQ}, \texttt{DPQ}, and \texttt{BDPQ}) and five out-of-distribution (OOD) tasks (\texttt{MPQ}, \texttt{BDMQ}, \texttt{BHMQ}, \texttt{BMPQ}, and \texttt{HMPQ}).
IND tasks evaluate the model's optimization capability on property combinations seen during training, while OOD tasks assess its zero-shot generalization to unseen property combinations.
%
%

\subsection{Evaluation Metrics}
\label{sec:experimental_setup:evaluation_metrics}
Following \gellmo, we evaluate optimization performance using three complementary metrics: \textbf{(1) Success Rate} ($\sr_\alpha$), which is the proportion of source molecules successfully optimized to improve all property objectives while maintaining Tanimoto similarity of at least $\alpha$ to the source molecule; \textbf{(2) Similarity} ($\similarity$) that is the average Tanimoto similarity between the optimized molecules and their corresponding source molecules; and \textbf{(3) Relative Improvement} ($\RI$) measuring the average relative change across the target properties with respect to the source molecules (Appendix~\ref{app:evaluation_metrics}).
Higher $\sr_\alpha$, $\similarity$ and $\RI$ are desirable, indicating better overall optimization performance.
%
%
%
For each source molecule $\source$, we select the optimized molecule $C$ as the candidate with the highest $\RI$ among those that improve all target properties and satisfy the similarity threshold $\alpha$.
If no such candidate exists, the optimization is considered unsuccessful.
%
%
%
%
%
%

\subsection{Baselines}
\label{sec:experimental_setup:baselines}
We evaluate \gellmomfg against the base model \gellmoPSix, and two categories of LLMs with different levels of chemistry-specific training: general-purpose LLMs and LLMs for chemistry which are trained on large-scale chemistry instruction tasks. 
We use \gellmoPSix to directly assess the improvement introduced by \framework while keeping the underlying model architecture fixed.
%
In terms of general-purpose LLMs, we select two of the largest and most capable open-weight models available to us at the time of evaluation, \qwen~\citep{qwen35} and \ministral~\citep{liu2026ministral}.
They serve as references for molecular optimization without adaptation to the chemistry domain.
We choose \llasmol~\citep{yu2024llasmol} and \biomedgpt~\citep{zuo2025biomedgpt} as LLMs for chemistry baselines due to their reported strong performance on \MOMO tasks.
%
%
For fair comparison, all baselines use identical inputs, inference settings (Section~\ref{sec:experimental_setup:inference}), and output processing procedures, with model and prompt details provided in Appendix~\ref{app:prompts}.
\subsection{Inference Settings}
\label{sec:experimental_setup:inference}
We evaluate all methods under both single-turn and multi-turn settings, denoted as ST and MT, respectively.
For baselines originally designed for ST optimization,
we extend them to the MT setting by applying iterative inference
with the same feedback vector and selection criterion as \framework.
This allows us to separate the contribution of \framework inference from \framework training.

Following the evaluation protocol of \gellmo,
we generate a maximum of 20 molecules throughout the optimization process for each source molecule.
%
%
For $T$-turn optimization,
we generate an equal number of
candidates at each turn.
%
%
%
%
%
To ensure a comprehensive evaluation,
we independently evaluate each method with $T\in\{1,2,3,4\}$ turns.
We report the results corresponding to the highest
\sr across the evaluated values of $T$ for each task.
The full results for each value of $T$ are provided in Appendix~\ref{app:results}.
\section{Results}
\label{sec:results}
\subsection{Performance on \MOMO tasks}
\label{sec:results:main}

\subsubsection{Overall comparison with baselines}
\label{sec:results:main:main}
\newcolumntype{T}{%
    r@{\extracolsep{0pt}\hspace{4pt}}%
    r@{\hspace{4pt}}%
    r%
}

\begin{table*}[t]
\centering
\setlength{\tabcolsep}{3pt}
\renewcommand{\arraystretch}{1.08}

\begin{threeparttable}
\caption{Overall Performance on IND Tasks}
\label{tab:main_ind}

\footnotesize
\begin{tabular*}{\textwidth}{
    @{}
    l
    @{\hspace{7pt}}
    c
    @{\extracolsep{\fill}\hspace{9pt}}
    T
    @{\extracolsep{\fill}\hspace{11pt}}
    T
    @{\extracolsep{\fill}\hspace{11pt}}
    T
    @{\extracolsep{\fill}\hspace{11pt}}
    T
    @{\extracolsep{\fill}\hspace{11pt}}
    T
    @{}
}

\toprule

\multirow{2}{*}{Model}
&
\multirow{2}{*}{Setting}
&
\multicolumn{3}{c}{\texttt{BDP}}
&
\multicolumn{3}{c}{\texttt{BDQ}}
&
\multicolumn{3}{c}{\texttt{BPQ}}
&
\multicolumn{3}{c}{\texttt{DPQ}}
&
\multicolumn{3}{c}{\texttt{BDPQ}}
\\

\cmidrule(lr){3-5}
\cmidrule(lr){6-8}
\cmidrule(lr){9-11}
\cmidrule(lr){12-14}
\cmidrule(lr){15-17}

&
&
$\sr_{0.5}^\uparrow$
&
$\similarity^\uparrow$
&
$\RI^\uparrow$
&
$\sr_{0.5}^\uparrow$
&
$\similarity^\uparrow$
&
$\RI^\uparrow$
&
$\sr_{0.5}^\uparrow$
&
$\similarity^\uparrow$
&
$\RI^\uparrow$
&
$\sr_{0.5}^\uparrow$
&
$\similarity^\uparrow$
&
$\RI^\uparrow$
&
$\sr_{0.5}^\uparrow$
&
$\similarity^\uparrow$
&
$\RI^\uparrow$
\\

\midrule

\rowcolor{gray!20}
\multicolumn{17}{c}{\bfseries General-purpose LLMs}
\\

\multirow{2}{*}{\qwen}
& ST
& 32.80 & \textbf{0.76} & 0.98
& 18.80 & \textbf{0.73} & 0.54
& 90.40 & \textbf{0.75} & 0.58
& 15.60 & \textbf{0.72} & 0.51
& 11.00 & \underline{0.75} & 0.54
\\

& MT
& 36.40 & \underline{0.72} & 0.95
& 21.60 & \underline{0.70} & 0.60
& 90.40 & \textbf{0.75} & 0.58
& 19.80 & \underline{0.69} & 0.36
& 12.20 & \textbf{0.76} & 0.51
\\

\multirow{2}{*}{\ministral}
& ST
& 27.60 & 0.71 & 0.74
& 24.00 & 0.67 & 0.45
& 64.80 & \underline{0.68} & 0.52
& 12.80 & 0.68 & 0.28
& 8.60 & 0.72 & 0.56
\\

& MT
& 37.60 & 0.68 & 0.70
& 24.60 & 0.66 & 0.69
& 66.00 & 0.65 & 0.65
& 13.60 & 0.68 & 0.36
& 11.60 & 0.70 & 0.47
\\

\rowcolor{gray!20}
\multicolumn{17}{c}{\bfseries LLMs for Chemistry}
\\

\multirow{2}{*}{\llasmol}
& ST
& 5.00 & 0.65 & 0.66
& 1.60 & 0.56 & 0.79
& 27.40 & 0.65 & 0.66
& 4.40 & 0.66 & 0.37
& 2.40 & 0.73 & 0.42
\\

& MT
& 6.60 & 0.62 & 0.96
& 1.60 & 0.56 & 0.79
& 30.00 & 0.62 & 0.72
& 5.00 & 0.63 & 0.37
& 3.60 & 0.64 & 0.48
\\

\multirow{2}{*}{\biomedgpt}
& ST
& 69.20 & 0.62 & 2.82
& \underline{79.00} & 0.62 & \underline{4.40}
& \textbf{97.20} & 0.61 & 1.23
& 47.80 & 0.63 & 1.58
& 38.00 & 0.62 & 2.03
\\

& MT
& 74.20 & 0.58 & 3.36
& \underline{79.00} & 0.62 & \underline{4.40}
& \textbf{97.20} & 0.61 & 1.23
& 47.80 & 0.63 & 1.58
& 43.00 & 0.59 & 1.88
\\

\rowcolor{gray!20}
\multicolumn{17}{c}{\bfseries \gellmo}
\\

\multirow{2}{*}{\gellmo}
& ST
& 66.00 & 0.62 & 2.77
& 65.00 & 0.64 & 3.88
& 84.40 & 0.60 & \underline{1.28}
& 43.80 & 0.61 & 1.84
& 39.00 & 0.60 & 2.26
\\

& MT
& 66.00 & 0.62 & 2.77
& 71.40 & 0.60 & 4.37
& 84.40 & 0.60 & \underline{1.28}
& 43.80 & 0.61 & 1.84
& 39.00 & 0.60 & 2.26
\\

\multirow{2}{*}{\gellmomfg}
& ST
& \underline{83.20} & 0.62 & \underline{3.62}
& 73.00 & 0.63 & 4.30
& 89.20 & 0.62 & 1.25
& \underline{48.40} & 0.63 & \underline{1.97}
& \underline{44.00} & 0.61 & \underline{2.50}
\\

& MT
& \textbf{84.20} & 0.61 & \textbf{4.24}
& \textbf{86.20} & 0.61 & \textbf{5.32}
& \underline{91.20} & 0.60 & \textbf{1.32}
& \textbf{54.60} & 0.60 & \textbf{2.49}
& \textbf{53.40} & 0.59 & \textbf{3.19}
\\

\bottomrule
\end{tabular*}

\begin{tablenotes}[flushleft]
\footnotesize
\item\hspace{-6pt}
ST denotes the original single-turn setting and MT denotes the multi-turn setting.
$\uparrow$ denotes that a higher value of the metric is desirable.
For each metric, the best performance is in \textbf{bold} and the second-best performance is \underline{underlined}.
\end{tablenotes}

\end{threeparttable}
\vspace{-8pt}
\end{table*}

%
Tables~\ref{tab:main_ind} and~\ref{tab:main_ood} present a comprehensive comparison between \gellmomfg and all the baselines on IND and OOD \MOMO tasks, respectively. 
Overall, \gellmomfg achieves the highest $\srZeroFive$ on 6 of the 10 tasks and highest \RI on 9 tasks.
Specifically, it achieves higher average $\srZeroFive$ ($81.50\%$ vs. $77.81\%$) and $\RI$ ($2.63$ vs. $2.08$) than the strongest baseline, \biomedgpt.
%
%
Given that the inference settings are consistent across all methods, 
the superior performance of \gellmomfg can be attributed to \framework's training design,
rather than the inference strategy alone.
\gellmomfg also maintains an average Sim of $0.60$, matching \gellmo ($0.60$) and \biomedgpt ($0.60$), indicating that stronger property optimization does not compromise average structural similarity. 
This further indicates that the similarity-sensitive reward assist the optimization process in preserving desired molecular structures.

%
%
%
%

\newcolumntype{T}{%
    r@{\extracolsep{0pt}\hspace{4pt}}%
    r@{\hspace{4pt}}%
    r%
}

\begin{table*}[t]
\centering
\setlength{\tabcolsep}{3pt}
\renewcommand{\arraystretch}{1.08}

\begin{threeparttable}
\caption{Overall Performance on OOD Tasks}
\label{tab:main_ood}

\footnotesize
\begin{tabular*}{\textwidth}{
    @{}
    l
    @{\hspace{7pt}}
    c
    @{\extracolsep{\fill}\hspace{9pt}}
    T
    @{\extracolsep{\fill}\hspace{11pt}}
    T
    @{\extracolsep{\fill}\hspace{11pt}}
    T
    @{\extracolsep{\fill}\hspace{11pt}}
    T
    @{\extracolsep{\fill}\hspace{11pt}}
    T
    @{}
}
\toprule

\multirow{2}{*}{Model}
&
\multirow{2}{*}{Setting}
&
\multicolumn{3}{c}{\texttt{MPQ}}
&
\multicolumn{3}{c}{\texttt{BDMQ}}
&
\multicolumn{3}{c}{\texttt{BHMQ}}
&
\multicolumn{3}{c}{\texttt{BMPQ}}
&
\multicolumn{3}{c}{\texttt{HMPQ}}
\\

\cmidrule(lr){3-5}
\cmidrule(lr){6-8}
\cmidrule(lr){9-11}
\cmidrule(lr){12-14}
\cmidrule(lr){15-17}

&
&
$\sr_{0.5}^\uparrow$
&
$\similarity^\uparrow$
&
$\RI^\uparrow$
&
$\sr_{0.5}^\uparrow$
&
$\similarity^\uparrow$
&
$\RI^\uparrow$
&
$\sr_{0.5}^\uparrow$
&
$\similarity^\uparrow$
&
$\RI^\uparrow$
&
$\sr_{0.5}^\uparrow$
&
$\similarity^\uparrow$
&
$\RI^\uparrow$
&
$\sr_{0.5}^\uparrow$
&
$\similarity^\uparrow$
&
$\RI^\uparrow$
\\

\midrule

\rowcolor{gray!20}
\multicolumn{17}{c}{\bfseries General-purpose LLMs}
\\

\multirow{2}{*}{\qwen}
& ST
& 75.80 & \textbf{0.69} & 0.41
& 17.00 & \textbf{0.76} & 0.41
& 72.03 & \textbf{0.73} & 1.03
& 75.39 & \textbf{0.71} & 0.47
& 81.25 & \textbf{0.73} & 0.54
\\

& MT
& 75.80 & \textbf{0.69} & 0.41
& 23.40 & \underline{0.71} & 0.43
& 77.12 & \underline{0.69} & 1.03
& 75.39 & \textbf{0.71} & 0.47
& 81.25 & \underline{0.69} & 0.68
\\

\multirow{2}{*}{\ministral}
& ST
& 47.80 & \underline{0.65} & 0.37
& 15.40 & 0.66 & 0.52
& 51.69 & 0.66 & 1.07
& 47.12 & \underline{0.65} & 0.47
& 34.38 & 0.67 & 0.59
\\

& MT
& 49.60 & 0.64 & 0.39
& 23.20 & 0.64 & 0.76
& 57.63 & 0.63 & 1.03
& 56.54 & \underline{0.65} & 0.49
& 40.62 & 0.63 & 0.69
\\

\rowcolor{gray!20}
\multicolumn{17}{c}{\bfseries LLMs for Chemistry}
\\

\multirow{2}{*}{\llasmol}
& ST
& 17.40 & 0.61 & 0.38
& 0.20 & 0.67 & 0.0
& 15.25 & 0.62 & 1.78
& 15.71 & 0.61 & 0.51
& 31.25 & 0.63 & 0.69
\\

& MT
& 18.20 & 0.60 & 0.41
& 0.80 & 0.58 & 0.68
& 20.34 & 0.63 & 1.02
& 18.32 & 0.64 & 0.50
& 31.25 & 0.63 & 0.69
\\

\multirow{2}{*}{\biomedgpt}
& ST
& \textbf{94.40} & 0.60 & 0.74
& 65.60 & 0.61 & 2.48
& 88.98 & 0.62 & 2.25
& \underline{91.10} & 0.60 & 0.87
& \underline{91.67} & 0.62 & 1.14
\\

& MT
& \textbf{94.40} & 0.60 & 0.74
& 65.60 & 0.61 & 2.48
& \textbf{91.53} & 0.59 & 2.53
& \textbf{91.62} & 0.60 & 0.92
& \textbf{93.75} & 0.59 & \textbf{1.64}
\\

\rowcolor{gray!20}
\multicolumn{17}{c}{\bfseries \gellmo}
\\

\multirow{2}{*}{\gellmo}
& ST
& 83.40 & 0.60 & \underline{0.76}
& 64.40 & 0.62 & 2.66
& 82.20 & 0.60 & \underline{2.81}
& 84.82 & 0.59 & \underline{0.96}
& 87.50 & 0.59 & 1.51
\\

& MT
& 83.40 & 0.60 & \underline{0.76}
& 65.60 & 0.59 & \underline{3.06}
& 82.20 & 0.60 & \underline{2.81}
& 84.82 & 0.59 & \underline{0.96}
& 87.50 & 0.59 & 1.51
\\

\multirow{2}{*}{\gellmomfg}
& ST
& \underline{88.40} & 0.60 & \textbf{0.78}
& \underline{80.20} & 0.62 & 2.77
& 80.51 & 0.60 & 2.74
& 86.91 & 0.62 & 0.92
& 88.54 & 0.60 & 1.37
\\

& MT
& \underline{88.40} & 0.60 & \textbf{0.78}
& \textbf{83.60} & 0.61 & \textbf{3.50}
& \underline{90.68} & 0.57 & \textbf{2.87}
& 89.01 & 0.60 & \textbf{0.99}
& \textbf{93.75} & 0.58 & \underline{1.62}
\\

\bottomrule
\end{tabular*}

\begin{tablenotes}[flushleft]
\footnotesize
\item\hspace{-6pt}
The metrics, notations and formatting have the same meanings
as those in Table \ref{tab:main_ind}.
\end{tablenotes}

\end{threeparttable}
\vspace{-8pt}
\end{table*}

\subsubsection{Comparison with \gellmo}
\label{sec:results:main:gellmo}
As shown in Tables~\ref{tab:main_ind} and~\ref{tab:main_ood}, \gellmomfg consistently outperforms its base model, \gellmo, across all IND and OOD tasks. 
Under the MT setting, \gellmomfg outperforms \gellmo across all IND and OOD tasks, achieving an average increase of $10.69\%$ in $\srZeroFive$ and 0.47 in \RI.
%
%
Moreover, the gap between ST and MT performance is substantially larger for \gellmomfg than for \gellmo.
Moving from ST to MT increases \gellmomfg's average $\srZeroFive$ by 5.27\% (compared with only 0.76 for \gellmo) and improves \RI on 9 tasks (compared with only 2 for \gellmo), while decreasing average $\similarity$ by only 0.02.
As both models share the same backbone and are evaluated under identical inference settings, the larger gains from ST to MT setting suggest that \framework's training better equips \gellmomfg to leverage intermediate molecules and their feedback to guide subsequent optimization.
These findings are also consistent with \framework's reward design, which promotes progressive improvement over previously generated molecules.
\newcolumntype{T}{%
    r@{\extracolsep{0pt}\hspace{4pt}}%
    r@{\hspace{4pt}}%
    r%
}

\begin{table*}[t]
\centering
\setlength{\tabcolsep}{3pt}
\renewcommand{\arraystretch}{1.08}

\begin{threeparttable}
\caption{Ablation Study}
\label{tab:ablation}

\fontsize{8.5pt}{9pt}\selectfont
\begin{tabular*}{\textwidth}{
    @{}
    l
    @{\hspace{4pt}}
    c
    @{\hspace{4pt}}
    T
    @{\extracolsep{\fill}\hspace{9pt}}
    T
    @{\extracolsep{\fill}\hspace{9pt}}
    T
    @{\extracolsep{\fill}\hspace{9pt}}
    T
    @{\extracolsep{\fill}\hspace{9pt}}
    T
    @{}
}
\toprule

\multirow{2}{*}{Model}
&
\multirow{2}{*}{Setting}
&
\multicolumn{3}{c}{\texttt{BDP}}
&
\multicolumn{3}{c}{\texttt{BDQ}}
&
\multicolumn{3}{c}{\texttt{BPQ}}
&
\multicolumn{3}{c}{\texttt{DPQ}}
&
\multicolumn{3}{c}{\texttt{BDPQ}}
\\

\cmidrule(lr){3-5}
\cmidrule(lr){6-8}
\cmidrule(lr){9-11}
\cmidrule(lr){12-14}
\cmidrule(lr){15-17}

& &
$\sr_{0.5}^\uparrow$ & $\similarity^\uparrow$ & $\RI^\uparrow$ &
$\sr_{0.5}^\uparrow$ & $\similarity^\uparrow$ & $\RI^\uparrow$ &
$\sr_{0.5}^\uparrow$ & $\similarity^\uparrow$ & $\RI^\uparrow$ &
$\sr_{0.5}^\uparrow$ & $\similarity^\uparrow$ & $\RI^\uparrow$ &
$\sr_{0.5}^\uparrow$ & $\similarity^\uparrow$ & $\RI^\uparrow$
\\

\midrule


\multirow{2}{*}{\gellmoBaseRL}
& ST
& 79.00 & \underline{0.61} & 3.67
& 70.60 & \textbf{0.63} & 4.94
& \underline{90.60} & \underline{0.60} & 1.31
& 47.20 & \underline{0.62} & \underline{2.00}
& 41.80 & \textbf{0.62} & 2.28
\\

& MT
& 79.40 & 0.58 & \underline{4.18}
& \underline{80.60} & 0.59 & \textbf{5.67}
& \underline{90.60} & \underline{0.60} & 1.31
& 47.20 & \underline{0.62} & \underline{2.00}
& \underline{49.20} & 0.59 & \underline{3.14}
\\

\multirow{2}{*}{\gellmomfg}
& ST
& \underline{83.20} & \textbf{0.62} & 3.62
& 73.00 & \textbf{0.63} & 4.30
& 89.20 & \textbf{0.62} & 1.25
& \underline{48.40} & \textbf{0.63} & 1.97
& 44.00 & \underline{0.61} & 2.50
\\

& MT
& \textbf{84.20} & \underline{0.61} & \textbf{4.24}
& \textbf{86.20} & \underline{0.61} & \underline{5.32}
& \textbf{91.20} & \underline{0.60} & \textbf{1.32}
& \textbf{54.60} & 0.60 & \textbf{2.49}
& \textbf{53.40} & 0.59 & \textbf{3.19}
\\

\midrule


& &
$\sr_{0.6}^\uparrow$ & $\similarity^\uparrow$ & $\RI^\uparrow$ &
$\sr_{0.6}^\uparrow$ & $\similarity^\uparrow$ & $\RI^\uparrow$ &
$\sr_{0.6}^\uparrow$ & $\similarity^\uparrow$ & $\RI^\uparrow$ &
$\sr_{0.6}^\uparrow$ & $\similarity^\uparrow$ & $\RI^\uparrow$ &
$\sr_{0.6}^\uparrow$ & $\similarity^\uparrow$ & $\RI^\uparrow$
\\

\midrule

\multirow{2}{*}{\gellmoBaseRL}
& ST
& 64.60 & \textbf{0.67} & 2.96
& 59.60 & \textbf{0.67} & 4.03
& \textbf{79.60} & \textbf{0.67} & 1.06
& 32.80 & \underline{0.67} & 1.58
& 28.00 & \textbf{0.69} & 1.25
\\

& MT
& 64.60 & \textbf{0.67} & \underline{3.17}
& \underline{64.00} & \textbf{0.67} & \underline{4.34}
& \textbf{79.60} & \textbf{0.67} & \underline{1.08}
& 32.80 & \underline{0.67} & \textbf{1.93}
& \underline{30.20} & \textbf{0.69} & \underline{1.90}
\\

\multirow{2}{*}{\gellmomfg}
& ST
& \underline{68.60} & \textbf{0.67} & 2.90
& 61.20 & \textbf{0.67} & 3.65
& \underline{78.60} & \textbf{0.67} & 1.03
& \underline{35.60} & \textbf{0.68} & 1.43
& 29.60 & \underline{0.68} & 1.65
\\

& MT
& \textbf{69.20} & \textbf{0.67} & \textbf{3.43}
& \textbf{69.40} & \textbf{0.67} & \textbf{4.56}
& \underline{78.60} & \textbf{0.67} & \textbf{1.09}
& \textbf{37.60} & \textbf{0.68} & \underline{1.80}
& \textbf{32.40} & \underline{0.68} & \textbf{2.00}
\\


\midrule

\multirow{2}{*}{Model}
&
\multirow{2}{*}{Setting}
&
\multicolumn{3}{c}{\texttt{MPQ}}
&
\multicolumn{3}{c}{\texttt{BDMQ}}
&
\multicolumn{3}{c}{\texttt{BHMQ}}
&
\multicolumn{3}{c}{\texttt{BMPQ}}
&
\multicolumn{3}{c}{\texttt{HMPQ}}
\\

\cmidrule(lr){3-5}
\cmidrule(lr){6-8}
\cmidrule(lr){9-11}
\cmidrule(lr){12-14}
\cmidrule(lr){15-17}

& &
$\sr_{0.5}^\uparrow$ & $\similarity^\uparrow$ & $\RI^\uparrow$ &
$\sr_{0.5}^\uparrow$ & $\similarity^\uparrow$ & $\RI^\uparrow$ &
$\sr_{0.5}^\uparrow$ & $\similarity^\uparrow$ & $\RI^\uparrow$ &
$\sr_{0.5}^\uparrow$ & $\similarity^\uparrow$ & $\RI^\uparrow$ &
$\sr_{0.5}^\uparrow$ & $\similarity^\uparrow$ & $\RI^\uparrow$
\\

\midrule


\multirow{2}{*}{\gellmoBaseRL}
& ST
& \textbf{88.80} & \underline{0.59} & \textbf{0.83}
& 69.80 & \textbf{0.62} & 3.01
& 88.14 & \underline{0.59} & \underline{2.87}
& \textbf{90.05} & \underline{0.61} & \textbf{0.99}
& 92.71 & \underline{0.59} & 1.44
\\

& MT
& \textbf{88.80} & \underline{0.59} & \textbf{0.83}
& \underline{81.60} & 0.59 & \textbf{3.77}
& \underline{88.98} & 0.58 & \textbf{3.12}
& \textbf{90.05} & \underline{0.61} & \textbf{0.99}
& \textbf{95.83} & 0.56 & \textbf{1.75}
\\

\multirow{2}{*}{\gellmomfg}
& ST
& \underline{88.40} & \textbf{0.60} & 0.78
& 80.20 & \textbf{0.62} & 2.77
& 80.51 & \textbf{0.60} & 2.74
& 86.91 & \textbf{0.62} & 0.92
& 88.54 & \textbf{0.60} & 1.37
\\

& MT
& \underline{88.40} & \textbf{0.60} & 0.78
& \textbf{83.60} & \underline{0.61} & \underline{3.50}
& \textbf{90.68} & 0.57 & \underline{2.87}
& \underline{89.01} & 0.60 & \textbf{0.99}
& \underline{93.75} & 0.58 & \underline{1.62}
\\

\midrule


& &
$\sr_{0.6}^\uparrow$ & $\similarity^\uparrow$ & $\RI^\uparrow$ &
$\sr_{0.6}^\uparrow$ & $\similarity^\uparrow$ & $\RI^\uparrow$ &
$\sr_{0.6}^\uparrow$ & $\similarity^\uparrow$ & $\RI^\uparrow$ &
$\sr_{0.6}^\uparrow$ & $\similarity^\uparrow$ & $\RI^\uparrow$ &
$\sr_{0.6}^\uparrow$ & $\similarity^\uparrow$ & $\RI^\uparrow$
\\

\midrule

\multirow{2}{*}{\gellmoBaseRL}
& ST
& \underline{71.20} & \textbf{0.66} & 0.72
& 58.00 & \underline{0.66} & 2.15
& \textbf{68.64} & \underline{0.67} & 2.25
& \underline{71.20} & \underline{0.67} & 0.83
& \textbf{72.92} & \underline{0.65} & \textbf{1.25}
\\

& MT
& \underline{71.20} & \textbf{0.66} & \underline{0.73}
& \underline{62.60} & \underline{0.66} & \textbf{2.69}
& \textbf{68.64} & \underline{0.67} & \underline{2.28}
& \underline{71.20} & \underline{0.67} & \textbf{0.90}
& \textbf{72.92} & \underline{0.65} & \textbf{1.25}
\\

\multirow{2}{*}{\gellmomfg}
& ST
& \textbf{75.20} & \textbf{0.66} & 0.67
& \textbf{68.80} & \textbf{0.67} & 1.97
& \underline{63.56} & \textbf{0.68} & \textbf{2.36}
& \underline{71.20} & \textbf{0.68} & 0.78
& \underline{71.88} & \textbf{0.67} & 1.13
\\

& MT
& \textbf{75.20} & \textbf{0.66} & \textbf{0.74}
& \textbf{68.80} & \textbf{0.67} & \underline{2.64}
& \textbf{68.64} & \textbf{0.68} & \textbf{2.36}
& \textbf{72.77} & \textbf{0.68} & \underline{0.86}
& \underline{71.88} & \textbf{0.67} & \underline{1.14}
\\

\bottomrule
\end{tabular*}

\begin{tablenotes}[flushleft]
\footnotesize
\item\hspace{-6pt}
The metrics, notations and formatting have the same meanings as those in Table \ref{tab:main_ind}.
\end{tablenotes}

\end{threeparttable}
\vspace{-8pt}
\end{table*}

\subsubsection{Comparison with other LLM baselines}
\label{sec:results:main:other}
Compared with \llasmol, \gellmomfg achieves vastly higher $\srZeroFive$ and $\RI$ across all tasks. 
Under MT, \llasmol averages $13.57\%$ and $0.66$ for $\srZeroFive$ and $\RI$, whereas \gellmomfg achieves $81.50\%$ and $2.63$, respectively.
This performance gap suggests that training on a broad collection of chemistry tasks alone may not develop strong molecular optimization capabilities or the ability to leverage property feedback and adapt optimization strategies accordingly.

\gellmomfg also substantially outperforms the general-purpose LLM baselines, achieving an average $\srZeroFive$ of $81.50\%$ under MT setting compared with $51.34\%$ for Qwen3.5-9B, the strongest general-purpose LLM baseline.
%
%
Moreover, \gellmomfg consistently achieves higher $\RI$ across all tasks, despite maintaining slightly lower but still substantial similarity to the source molecules. 
These results reinforce the importance of chemistry-specific or even task-specific training for developing strong \MOMO capabilities.
\subsection{Ablation Study}
\label{sec:ablation_study}
To investigate the contribution of MT feedback-guided training beyond single-turn RL training, we compare \gellmomfg with \gellmoBaseRL.
Both models are initialized from \gellmo and trained using GRPO under identical settings.
However, \gellmoBaseRL uses ST training ($T$ $=$ $1$) without intermediate feedback.
Because there are no intermediate molecules in this setting,
we only maximize $R_{\mathrm{base}}(\cdot)$ (Eq.~\ref{eq:reward_definition}) during \gellmoBaseRL's training.
To better study the differences between \gellmomfg and \gellmoBaseRL in terms of similarity-constrained optimization, we evaluate \sr under two different similarity thresholds:
0.5 and 0.6, denoted as $\srZeroFive$ and $\srZeroSix$, respectively.

%
%
%

%

\begin{figure*}[htb]
    \centering
    \includegraphics[width=\linewidth]{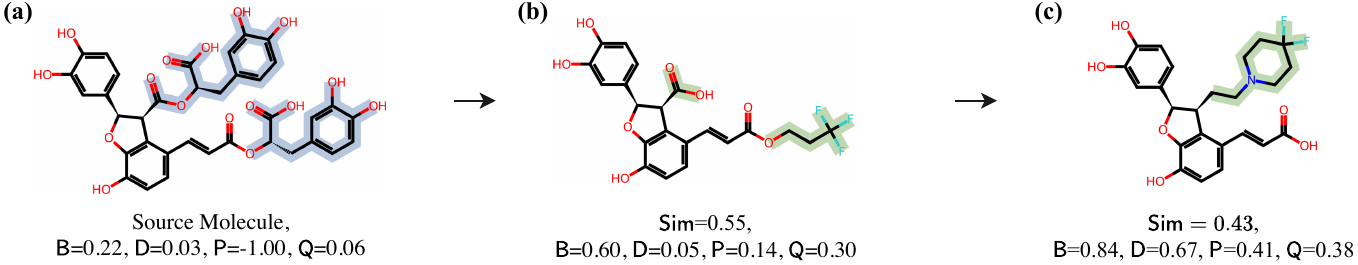}
    \caption{An optimization case on \texttt{BDPQ} with $T=2$. \textbf{(a)} The source molecule. \textbf{(b)} The best optimized molecule at turn $t = 1$. \textbf{(c)} The best optimized molecule at turn $t = 2$. The modified fragments are highlighted.}
    \label{fig:case}
\end{figure*}

\subsubsection{Overall comparison of \gellmomfg and \gellmoBaseRL}
As shown in Table~\ref{tab:ablation}, ST RL training alone leads to notable improvements in MOMO performance. 
Under ST inference, \gellmoBaseRL achieves a higher average $\srZeroFive$ than the base model \gellmo ($75.87\%$ vs. $70.05\%$).
Given that both models share the same backbone and inference setting, these results highlight the effectiveness of our reward function in guiding RL training.
Specifically, as \gellmoBaseRL was trained with ST setting, it indicates that the $R_{\mathrm{base}}$ 
component 
can provide a strong learning signal for \MOMO.
%
%

Building on this improvement, \gellmomfg further outperforms \gellmoBaseRL under MT inference, achieving higher average $\srZeroFive$ ($81.50\%$ vs. $79.23\%$) and $\srZeroSix$ ($64.45\%$ vs. $61.78\%$).
%
%
Moreover, moving from ST to MT increases $\srZeroFive$ and $\srZeroSix$ by approximately $5.27$ and $2.03\%$ for \gellmomfg, compared with smaller gains of $3.36$ and $1.12$ for \gellmoBaseRL, respectively.
As both models were trained with GRPO and follow identical inference setting, these larger gains highlight the additional benefits of \framework's multi-turn training, 
which incorporates intermediate feedback and the $R_{\mathrm{best}}$ reward component to encourage progressive improvements.
Together, these findings suggest that while $R_{\mathrm{base}}$ provides a useful foundation for RL-based molecular optimization, \framework's multi-turn training design further enhances the model's ability to leverage feedback and progressively refine molecules across turns.

%
%

\subsubsection{
Performance under different similarity thresholds
}
\label{sec:ablation:sr06}

The advantage of \framework becomes more pronounced under a more stringent similarity threshold $\alpha=0.6$ as demonstrated in Table~\ref{tab:ablation}. 
In the ST setting, \gellmomfg improves over \gellmoBaseRL by only $0.37\%$ on average at $\srZeroFive$, but by $1.77\%$ at $\srZeroSix$. 
Under the MT setting, the corresponding advantage increases from approximately $2.28\%$ at $\srZeroFive$ to $2.67\%$ at $\srZeroSix$. 
Both models use $\mathcal{S}(\cdot)$ (Eq.~\ref{eq:reward_definition}) in their reward function to encourage structural similarity, thus the larger performance gains under stricter similarity constraints suggest that \framework's multi-turn training enables the \gellmomfg
to better leverage intermediate feedback to adjust its optimization strategies.
The higher success rates also reflect \gellmomfg's ability to better balance property improvement and structural similarity preservation.

\subsection{Case Study}
\label{sec:results:case_study}

Figure~\ref{fig:case} shows a successful two-turn optimization by \gellmomfg on four properties: \texttt{B}, \texttt{D}, \texttt{P}, and \texttt{Q}.
%
%
In this optimization, the first turn substantially improves \texttt{B}, \texttt{P}, and \texttt{Q}, while the second turn focuses on improving \texttt{D}.
The source molecule (Figure~\ref{fig:case}a) 
contains multiple polyphenolic esters that contribute to 
low lipophilicity and high molecular weight,
both of which reduce \texttt{B} and \texttt{Q}.
In the first turn, \gellmomfg reduces polarity by hydrolyzing one phenolic ester and replacing another with a 1,1,1-trifluoropropyl ester.
These modifications also reduce the molecular weight from 718 to 461,
which improves \texttt{Q} by aligning better with molecular weight guidelines for drug-like molecules, 
such as Lipinski's Rule of Five~\cite{lipinski}.
However, the changes introduced in this turn change very little to improve the molecule's predicted DRD2 activity, increasing from 0.03 to 0.05.

Based on this feedback, \gellmomfg prioritizes \texttt{D} in the second turn,
replacing the free dihydrobenzofuran carboxylic acid with a 4,4-difluoropiperidinyl group and hydrolyzing the 1,1,1-trifluoropropyl ester.
This produces a substantial improvement on \texttt{D}, while also further improving \texttt{B}, \texttt{P}, and \texttt{Q}.
The improvement on \texttt{D} may be driven in part by the introduction of a cyclic amine,
which helps the molecule better resemble patterns in established drugs targeting DRD2.
Taken together, this multi-turn optimization example
illustrates how \gellmomfg shifts priorities across turns based on feedback,
resulting in higher-quality candidate molecules after multiple turns.


\vspace{-2pt}
\section{Conclusion}
\label{sec:conclusion}
\vspace{-2pt}

In this work, we introduce \framework for molecular optimization, a multi-turn RL framework that trains LLMs to iteratively optimize molecules using previously generated candidates and their property feedback.
By training a SOTA molecular optimization LLM, \gellmo, with \framework, 
we develop \gellmomfg.
Across 10 \MOMO tasks, \gellmomfg outperforms \gellmo and achieves competitive performance against existing baselines, improving molecular properties while preserving structural similarity.
Extensive analyses demonstrate \framework's effectiveness in leveraging multi-turn feedback for progressive molecular optimization.
Together, these findings highlight the advantages of \framework in training LLMs to adapt their molecular optimization strategies across turns.
%


\section*{Acknowledgments}
This project was made possible, in part, by support from the National Science Foundation grant no. IIS-2435819 (X.N.), the National Library of Medicine grant no. 1R01LM014385 (X.N.), and the Sanofi iDEA-TECH Awards North America (X.N.). Any opinions, findings, conclusions or recommendations expressed in this paper are those of the authors and do not necessarily reflect the views of the funding agency.  The authors would like to thank Benjamin Burns for their proofreading of the manuscript.


\section*{Limitations}

Despite the strong performance of \framework in multi-objective molecular optimization, several limitations remain. 
(1) Our experiments are conducted on \mumoinstruct, which includes \MOMO tasks involving six target properties. 
Although \gellmomfg demonstrates strong generalization to out-of-distribution (OOD) tasks with unseen property combinations, its effectiveness on additional property objectives remains unexplored. 
(2) \framework relies on property oracles to provide intermediate property feedback and compute rewards for generated molecules. 
The reliability of this feedback depends on the accuracy of the property oracles, and the improvements observed in our evaluations have not been experimentally validated. 
(3) During training, the reward function uses the target molecule provided by \mumoinstruct to normalize property improvements and structural similarity. 
This requirement may limit the applicability of the current training framework to datasets without corresponding target molecules. 
(4) We apply \framework to \gellmo to develop \gellmomfg. Although the results demonstrate the effectiveness of \framework on this model, further evaluations on other molecular optimization LLMs are needed to establish the generalizability of our framework.

\section*{Ethical considerations}

\framework for molecular optimization aims to improve multiple property objectives through iterative generation guided by intermediate property feedback.
Although our framework is developed for molecular optimization rather than harmful applications, its ability to generate and optimize candidate molecules may also be used to produce potentially unsafe compounds. 
Moreover, improvements in target property objectives and structural similarity to the source molecule do not necessarily indicate that the generated molecules are safe for practical use. 
We therefore emphasize that molecules generated by \framework should undergo careful safety evaluation by expert chemists before being considered for synthesis or experimental testing.
Any practical application of the framework and its generated molecules should also adhere to relevant safety regulations, ethical guidelines, and established professional practices.

\bibliography{latex/custom}

\clearpage
\appendix
\setcounter{table}{0}
\renewcommand{\thetable}{\Alph{section}.\arabic{table}}
\counterwithin{table}{section}

\section{Appendix}

\subsection{Related Work}
\label{app:related_work}

\paragraph{Molecular optimization}
Computational methods for molecular optimization has been explored through a variety of paradigms, including molecule string translation and graph-based generation.
Chemformer \cite{Irwin_2022} and Prompt-MolOpt \cite{Wu2024LeveragingLM} used transformer models to translate input molecule strings into optimized molecules.
Modof \cite{chen2021deep} and MIMOSA \cite{fu2021mimosa} instead formulated molecular optimization as a graph modification problem and designed graph neural network-based methods to edit substructures. 
However, these methods are developed for specific objectives and may have limited generalizability across diverse optimization tasks.
More recently, LLMs have shown the potential for molecular optimization by treating molecule string as a chemical language.
For example, ChatDrug \cite{liu2023chatgpt} and Re$^2$DF \cite{le2024agentdrug} designed inference-time methods to iteratively refine molecules based on LLM dialogue with external domain feedback.
DrugAssist \cite{ye2025drugassist} and GeLLM$^3$O \cite{dey-etal-2025-gellm3o} further instruction-tuned general-purpose LLMs specifically for molecular optimization.
BioMedGPT-Mol \cite{zuo2025biomedgpt} incorporated molecular optimization as one of multiple chemical tasks in a foundation chemical language model.
Building on these ideas, our work finetunes a molecular optimization LLM via reinforcement learning (RL) so that it can be applied multiple times to generate progressively improved molecules.

\subsection{Selection Criterion}
\label{app:selection}
$\operatorname{Advance}(\cdot)$ ranks the intermediate candidates according to the criterion in Table~\ref{tab:next_selection}, listed in descending order of priority.
Each criterion is considered only when higher-priority criteria cannot identify a unique candidate.
\begin{table}[htbp]
\centering
\begin{threeparttable}
\caption{Selection criteria of $\operatorname{Advance}(\cdot)$ in descending order of priority.}
\label{tab:next_selection}
\small
\begin{tabular}{cl}
\toprule
Priority & Selection Criterion \\
\midrule
1 & Chemical validity \\
2 & Similarity to $\source$ $\geq \delta$ \\
3 & Improvement on all properties in $\propertyset$ \\
4 & Number of improved properties \\
5 & Similarity to $\source$ $\geq \delta_{\mathrm{h}}$ \\
6 & Relative property improvement \\
7 & Similarity to $\source$ \\
\bottomrule
\end{tabular}
\end{threeparttable}
\end{table}

\subsection{Reward Function}
\label{app:reward_function}
\paragraph{Definition of $\mathcal{G}(\cdot)$.}

For each source molecule $\source$,
we use the reference molecule $C_{\mathrm{ref}}$ provided by \mumoinstruct to normalize the improvement in each property $p\in \propertyset$:
\begin{equation}
\mathcal{G}(C_t\mid p)
=
\operatorname{clip}
\left(
\frac{
d_p \Delta y_p(C_t, \source)
}{
d_p \Delta y_p(C_{\mathrm{target}}, \source)
},
-\kappa, \kappa
\right),
\label{app:eq:property_improv}
\end{equation}
where $\Delta y_p(A,B)=y_p(A)-y_p(B)$ denotes the difference in property $p$ between molecules $A$ and $B$, $d_p\in\{-1,1\}$ specifies the desired optimization direction, and $\kappa$ is the clipping threshold.

We compute the average normalized improvement across all property objectives:
\begin{equation}
\mathcal{G}(C_t)
=
\frac{1}{|\propertyset|}
\sum_{p \in \propertyset}
\mathcal{G}(C_t\mid p).
\end{equation}
%
By using $C_{\mathrm{ref}}$ for normalization,
property improvements are scaled to balance the contributions of different property objectives.

\paragraph{Definition of $\mathcal{S}(\cdot)$.}
We similarly normalize the structural similarity $C_{\mathrm{ref}}$:
%
\begin{equation}
\mathcal{S}(C_t) =
\begin{cases}
1,
& \operatorname{Sim}(C_t,\source)
  \geq \delta_{\mathrm{h}}, \\[3pt]
\displaystyle
\frac{
\operatorname{Sim}(C_t,\source)
}{
\operatorname{Sim}(C_{\mathrm{ref}},\source)
},
& \text{otherwise},
\end{cases}
\label{eq:similarity_reward_details}
\end{equation}
where $\operatorname{Sim}(A,B)\in[0,1]$ denotes the Tanimoto similarity between molecules $A$ and $B$. 
The threshold-based design encourages structural similarity while preventing it from dominating the optimization objective.

\paragraph{Definition of $\lambda(\cdot)$.}
The scaling factor $\lambda_t$ provides an additional bonus when $C_t$
improves all target properties over $\source$, 
as indicated by $A(C_t)
=
\mathbb{I}
\left[
d_p\Delta y_p(C_t,\source)>0,
\ \forall p\in\propertyset
\right]$.
%
%
We then define
\begin{equation}
\lambda(C_t)
=
1 + A(C_t)\,\operatorname{Sim}(C_t,\source).
\end{equation}
Thus, 
$\lambda(\cdot)>1$ only when $C_t$ simultaneously improves all properties, 
with a larger $\lambda(\cdot)$ for higher similarity.
%
%

\paragraph{Validity Penalties.}
An output is considered invalid if it does not contain exactly one tagged SMILES string, fails RDKit parsing or kekulization, or duplicates the source molecule or any previously generated molecule.
We assign a reward of $-2$ when $C_t$ is invalid.
For valid molecules,
we set the minimum reward to $-1$ to prevent them from receiving penalties comparable to those for invalid generations, 
despite suboptimal property improvements.


\newcommand{\bestTurn}[3]{%
    \cellcolor{yellow!18}#1 &
    \cellcolor{yellow!18}#2 &
    \cellcolor{yellow!18}#3%
}

\newcolumntype{T}{%
    r@{\extracolsep{0pt}\hspace{4pt}}%
    r@{\hspace{4pt}}%
    r%
}

\begin{table*}[t]
\centering
\setlength{\tabcolsep}{3pt}
\renewcommand{\arraystretch}{1.08}

\begin{threeparttable}
\caption{Overall Performance on IND Tasks Across Turn Settings}
\label{tab:all_turns_ind}

\footnotesize
\begin{tabular*}{\textwidth}{
    @{}
    l
    @{\hspace{7pt}}
    c
    @{\extracolsep{\fill}\hspace{9pt}}
    T
    @{\extracolsep{\fill}\hspace{11pt}}
    T
    @{\extracolsep{\fill}\hspace{11pt}}
    T
    @{\extracolsep{\fill}\hspace{11pt}}
    T
    @{\extracolsep{\fill}\hspace{11pt}}
    T
    @{}
}
\toprule

\multirow{2}{*}{Model}
&
\multirow{2}{*}{$T$}
&
\multicolumn{3}{c}{\texttt{BDP}}
&
\multicolumn{3}{c}{\texttt{BDQ}}
&
\multicolumn{3}{c}{\texttt{BPQ}}
&
\multicolumn{3}{c}{\texttt{DPQ}}
&
\multicolumn{3}{c}{\texttt{BDPQ}}
\\

\cmidrule(lr){3-5}
\cmidrule(lr){6-8}
\cmidrule(lr){9-11}
\cmidrule(lr){12-14}
\cmidrule(lr){15-17}

& &
$\sr_{0.5}^\uparrow$ & $\similarity^\uparrow$ & $\RI^\uparrow$ &
$\sr_{0.5}^\uparrow$ & $\similarity^\uparrow$ & $\RI^\uparrow$ &
$\sr_{0.5}^\uparrow$ & $\similarity^\uparrow$ & $\RI^\uparrow$ &
$\sr_{0.5}^\uparrow$ & $\similarity^\uparrow$ & $\RI^\uparrow$ &
$\sr_{0.5}^\uparrow$ & $\similarity^\uparrow$ & $\RI^\uparrow$
\\

\midrule

\rowcolor{gray!20}
\multicolumn{17}{c}{\bfseries General-purpose LLMs}
\\

\multirow{4}{*}{\qwen}
 & 1 & 32.80 & \textbf{0.76} & 0.98 & 18.80 & \textbf{0.73} & 0.54 & \bestTurn{90.40}{\textbf{0.75}}{0.58} & 15.60 & \textbf{0.72} & 0.51 & 11.00 & \underline{0.75} & 0.54 \\
 & 2 & \bestTurn{36.40}{\underline{0.72}}{0.95} & \bestTurn{21.60}{0.70}{0.60} & 88.60 & \underline{0.71} & 0.65 & \bestTurn{19.80}{0.69}{0.36} & \bestTurn{12.20}{\textbf{0.76}}{0.51} \\
 & 3 & 35.40 & 0.71 & 0.97 & 19.80 & 0.70 & 0.44 & 81.80 & \underline{0.71} & 0.61 & 16.60 & \underline{0.70} & 0.35 & 11.60 & 0.72 & 0.51 \\
 & 4 & 34.60 & 0.70 & 1.04 & 19.80 & \underline{0.71} & 0.51 & 75.80 & \underline{0.71} & 0.62 & 15.40 & 0.69 & 0.33 & 11.80 & 0.71 & 0.60 \\

 \addlinespace[3pt]

\multirow{4}{*}{\ministral}
 & 1 & 27.60 & 0.71 & 0.74 & 24.00 & 0.67 & 0.45 & 64.80 & 0.68 & 0.52 & 12.80 & 0.68 & 0.28 & 8.60 & 0.72 & 0.56 \\
 & 2 & 33.60 & 0.68 & 0.81 & \bestTurn{24.60}{0.66}{0.69} & \bestTurn{66.00}{0.65}{0.65} & \bestTurn{13.60}{0.68}{0.36} & \bestTurn{11.60}{0.70}{0.47} \\
 & 3 & 36.60 & 0.67 & 0.84 & 23.75 & 0.67 & 0.62 & 60.60 & 0.65 & 0.63 & 12.00 & 0.67 & 0.35 & 11.20 & 0.68 & 0.53 \\
 & 4 & \bestTurn{37.60}{0.68}{0.70} & 23.40 & 0.66 & 0.65 & 57.40 & 0.65 & 0.63 & 12.00 & 0.67 & 0.32 & 11.00 & 0.68 & 0.56 \\

\rowcolor{gray!20}
\multicolumn{17}{c}{\bfseries LLMs for Chemistry}
\\

\multirow{4}{*}{\llasmol}
 & 1 & 5.00 & 0.65 & 0.66 & \bestTurn{1.60}{0.56}{0.79} & 27.40 & 0.65 & 0.66 & 4.40 & 0.66 & 0.37 & 2.40 & 0.73 & 0.42 \\
 & 2 & 6.40 & 0.64 & 0.81 & 0.80 & 0.54 & 0.66 & \bestTurn{30.00}{0.62}{0.72} & \bestTurn{5.00}{0.63}{0.37} & 3.60 & 0.67 & 0.42 \\
 & 3 & 6.40 & 0.62 & 1.04 & 0.80 & 0.53 & 0.68 & 28.78 & 0.63 & 0.65 & \bestTurn{5.00}{0.63}{0.37} & \bestTurn{3.60}{0.64}{0.48} \\
 & 4 & \bestTurn{6.60}{0.62}{0.96} & 0.80 & 0.53 & 0.67 & 25.60 & 0.63 & 0.66 & 3.80 & 0.67 & 0.28 & 2.60 & 0.65 & 0.48 \\

\addlinespace[3pt]

\multirow{4}{*}{\biomedgpt}
 & 1 & 69.20 & 0.62 & 2.82 & \bestTurn{79.00}{0.62}{4.40} & \bestTurn{\textbf{97.20}}{0.61}{1.23} & \bestTurn{47.80}{0.63}{1.58} & 38.00 & 0.62 & 2.03 \\
 & 2 & 73.80 & 0.59 & 3.10 & 63.80 & 0.59 & 4.20 & \underline{96.60} & 0.59 & 1.26 & 43.80 & 0.61 & 2.07 & \bestTurn{43.00}{0.59}{1.88} \\
 & 3 & \bestTurn{74.20}{0.58}{3.36} & 60.40 & 0.60 & 4.31 & 95.40 & 0.59 & 1.25 & 41.80 & 0.61 & 2.05 & 42.00 & 0.59 & 1.91 \\
 & 4 & 73.60 & 0.58 & 3.28 & 58.40 & 0.59 & 4.46 & 94.80 & 0.58 & 1.24 & 41.40 & 0.60 & 2.04 & 42.00 & 0.59 & 2.06 \\

\rowcolor{gray!20}
\multicolumn{17}{c}{\bfseries \gellmo}
\\

\multirow{4}{*}{\gellmo}
 & 1 & \bestTurn{66.00}{0.62}{2.77} & 65.00 & 0.64 & 3.88 & \bestTurn{84.40}{0.60}{1.28} & \bestTurn{43.80}{0.61}{1.84} & \bestTurn{39.00}{0.60}{2.26} \\
 & 2 & 63.60 & 0.59 & 3.10 & 71.40 & 0.60 & 4.18 & 76.40 & 0.58 & 1.28 & 41.00 & 0.58 & 2.20 & 37.20 & 0.58 & 3.00 \\
 & 3 & 62.60 & 0.57 & 3.37 & \bestTurn{71.40}{0.60}{4.37} & 75.60 & 0.58 & 1.26 & 42.60 & 0.58 & 2.20 & 34.60 & 0.57 & \textbf{3.29} \\
 & 4 & 61.20 & 0.57 & 3.21 & 70.80 & 0.59 & 4.39 & 75.00 & 0.58 & 1.23 & 40.60 & 0.57 & 2.47 & 33.40 & 0.57 & \underline{3.22} \\

 \addlinespace[3pt]

\multirow{4}{*}{\gellmoBaseRL}
 & 1 & 79.00 & 0.61 & 3.67 & 70.60 & 0.63 & 4.94 & \bestTurn{90.60}{0.60}{1.31} & \bestTurn{47.20}{0.62}{2.00} & 41.80 & 0.62 & 2.28 \\
 & 2 & \bestTurn{79.40}{0.58}{4.18} & 79.40 & 0.60 & \textbf{5.88} & 89.00 & 0.58 & \underline{1.35} & 46.80 & 0.59 & \underline{2.77} & \bestTurn{49.20}{0.59}{3.14} \\
 & 3 & 78.80 & 0.57 & \textbf{4.34} & 80.40 & 0.58 & \underline{5.87} & 88.40 & 0.57 & \textbf{1.36} & 44.80 & 0.58 & \textbf{2.89} & 48.40 & 0.58 & 3.09 \\
 & 4 & 77.40 & 0.57 & \underline{4.30} & \bestTurn{80.60}{0.59}{5.67} & 87.40 & 0.57 & \textbf{1.36} & 44.40 & 0.58 & 2.68 & 47.20 & 0.58 & 3.07 \\

 \addlinespace[3pt]

\multirow{4}{*}{\gellmomfg}
 & 1 & \underline{83.20} & 0.62 & 3.62 & 73.00 & 0.63 & 4.30 & 89.20 & 0.62 & 1.25 & 48.40 & 0.63 & 1.97 & 44.00 & 0.61 & 2.50 \\
 & 2 & \bestTurn{\textbf{84.20}}{0.61}{4.24} & \underline{85.60} & 0.61 & 5.22 & \bestTurn{91.20}{0.60}{1.32} & 54.20 & 0.61 & 2.55 & 51.40 & 0.59 & 2.84 \\
 & 3 & \underline{83.20} & 0.61 & 3.90 & 84.60 & 0.61 & 5.40 & 89.60 & 0.59 & 1.34 & \underline{54.40} & 0.60 & 2.47 & \bestTurn{\textbf{53.40}}{0.59}{3.19} \\
 & 4 & 83.00 & 0.61 & 3.92 & \bestTurn{\textbf{86.20}}{0.61}{5.32} & 89.40 & 0.60 & 1.32 & \bestTurn{\textbf{54.60}}{0.60}{2.49} & \underline{51.80} & 0.59 & 3.16 \\

\bottomrule
\end{tabular*}
\begin{tablenotes}[flushleft]
\footnotesize
\item\hspace{-6pt}
The results with optimal $T$ for each model and task is highlighted in yellow.
Other metrics, notations, and formatting follow
Table \ref{tab:main_ind}.
\end{tablenotes}
\end{threeparttable}
\vspace{-8pt}
\end{table*}


\providecommand{\bestTurn}[3]{%
    \cellcolor{yellow!18}#1 &
    \cellcolor{yellow!18}#2 &
    \cellcolor{yellow!18}#3%
}

\newcolumntype{T}{%
    r@{\extracolsep{0pt}\hspace{4pt}}%
    r@{\hspace{4pt}}%
    r%
}

\begin{table*}[t]
\centering
\setlength{\tabcolsep}{3pt}
\renewcommand{\arraystretch}{1.08}

\begin{threeparttable}
\caption{Overall Performance on OOD Tasks Across Turn Settings}
\label{tab:all_turns_ood}

\footnotesize
\begin{tabular*}{\textwidth}{
    @{}
    l
    @{\hspace{7pt}}
    c
    @{\extracolsep{\fill}\hspace{9pt}}
    T
    @{\extracolsep{\fill}\hspace{11pt}}
    T
    @{\extracolsep{\fill}\hspace{11pt}}
    T
    @{\extracolsep{\fill}\hspace{11pt}}
    T
    @{\extracolsep{\fill}\hspace{11pt}}
    T
    @{}
}
\toprule

\multirow{2}{*}{Model}
&
\multirow{2}{*}{$T$}
&
\multicolumn{3}{c}{\texttt{MPQ}}
&
\multicolumn{3}{c}{\texttt{BDMQ}}
&
\multicolumn{3}{c}{\texttt{BHMQ}}
&
\multicolumn{3}{c}{\texttt{BMPQ}}
&
\multicolumn{3}{c}{\texttt{HMPQ}}
\\

\cmidrule(lr){3-5}
\cmidrule(lr){6-8}
\cmidrule(lr){9-11}
\cmidrule(lr){12-14}
\cmidrule(lr){15-17}

& &
$\sr_{0.5}^\uparrow$ & $\similarity^\uparrow$ & $\RI^\uparrow$ &
$\sr_{0.5}^\uparrow$ & $\similarity^\uparrow$ & $\RI^\uparrow$ &
$\sr_{0.5}^\uparrow$ & $\similarity^\uparrow$ & $\RI^\uparrow$ &
$\sr_{0.5}^\uparrow$ & $\similarity^\uparrow$ & $\RI^\uparrow$ &
$\sr_{0.5}^\uparrow$ & $\similarity^\uparrow$ & $\RI^\uparrow$
\\

\midrule

\rowcolor{gray!20}
\multicolumn{17}{c}{\bfseries General-purpose LLMs}
\\

\multirow{4}{*}{\qwen}
 & 1 & \bestTurn{75.80}{\textbf{0.69}}{0.41} & 17.00 & \textbf{0.76} & 0.41 & 72.03 & \textbf{0.73} & 1.03 & \bestTurn{75.39}{\textbf{0.71}}{0.47} & 81.25 & \textbf{0.73}& 0.54 \\
 & 2 & 68.00 & \underline{0.66} & 0.46 & \bestTurn{23.40}{\underline{0.71}}{0.43} & \bestTurn{77.12}{0.69}{1.03} & 73.82 & \underline{0.67} & 0.54 & \bestTurn{81.25}{\underline{0.69}}{0.68} \\
 & 3 & 63.20 & \underline{0.66} & 0.43 & 21.20 & 0.69 & 0.41 & 73.73 & 0.69 & 1.03 & 62.83 & 0.66 & 0.55 & 72.92 & 0.68 & 0.73 \\
 & 4 & 60.20 & \underline{0.66} & 0.45 & 21.40 & 0.69 & 0.42 & 70.34 & \underline{0.70} & 0.98 & 59.16 & \underline{0.67} & 0.52 & 70.83 & 0.67 & 0.74 \\

\addlinespace[3pt]

\multirow{4}{*}{\ministral}
 & 1 & 47.80 & 0.65 & 0.37 & 15.40 & 0.66 & 0.52 & 51.69 & 0.66 & 1.07 & 47.12 & 0.65 & 0.47 & 34.38 & 0.67 & 0.59 \\
 & 2 & \bestTurn{49.60}{0.64}{0.39} & \bestTurn{23.20}{0.64}{0.76} & \bestTurn{57.63}{0.63}{1.03} & \bestTurn{56.54}{0.65}{0.49} & \bestTurn{40.62}{0.63}{0.69} \\
 & 3 & 43.00 & 0.63 & 0.38 & 18.60 & 0.65 & 0.62 & 46.61 & 0.63 & 0.81 & 52.36 & 0.66 & 0.44 & 40.62 & 0.64 & 0.58 \\
 & 4 & 42.40 & 0.63 & 0.38 & 16.00 & 0.65 & 0.69 & 38.98 & 0.63 & 0.89 & 50.26 & 0.65 & 0.43 & 35.42 & 0.66 & 0.54 \\

\rowcolor{gray!20}
\multicolumn{17}{c}{\bfseries LLMs for Chemistry}
\\

\multirow{4}{*}{\llasmol}
 & 1 & 17.40 & 0.61 & 0.38 & 0.20 & 0.67 & 0.00 & 15.25 & 0.62 & 1.78 & 15.71 & 0.61 & 0.51 & \bestTurn{31.25}{0.63}{0.69} \\
 & 2 & \bestTurn{18.20}{0.60}{0.41} & 0.60 & 0.59 & 0.33 & \bestTurn{20.34}{0.63}{1.02} & \bestTurn{18.32}{0.64}{0.50} & 25.00 & 0.61 & 0.65 \\
 & 3 & 17.00 & 0.60 & 0.39 & \bestTurn{0.80}{0.58}{0.68} & 15.25 & 0.63 & 1.38 & 16.23 & 0.63 & 0.49 & 29.17 & 0.62 & 0.70 \\
 & 4 & 15.80 & 0.61 & 0.36 & 0.60 & 0.63 & 0.55 & 12.71 & 0.60 & 1.14 & 13.61 & 0.63 & 0.51 & 21.88 & 0.61 & 0.71 \\

\addlinespace[3pt]

\multirow{4}{*}{\biomedgpt}
 & 1 & \bestTurn{\textbf{94.40}}{0.60}{0.74} & \bestTurn{65.60}{0.61}{2.48} & 88.98 & 0.62 & 2.25 & \underline{91.10} & 0.60 & 0.87 & 91.67 & 0.62 & 1.14 \\
 & 2 & 81.00 & 0.59 & 0.81 & 57.20 & 0.60 & 2.64 & \bestTurn{\textbf{91.53}}{0.59}{2.53} & \bestTurn{\textbf{91.62}}{0.60}{0.92} & \bestTurn{93.75}{0.59}{1.64} \\
 & 3 & 79.00 & 0.59 & 0.80 & 54.60 & 0.59 & 2.65 & 89.83 & 0.58 & 2.60 & 88.48 & 0.60 & 0.94 & 86.46 & 0.58 & 1.42 \\
 & 4 & 78.20 & 0.59 & 0.80 & 52.20 & 0.59 & 2.79 & 87.29 & 0.59 & 2.42 & 84.82 & 0.59 & 0.92 & 86.46 & 0.58 & 1.47 \\

\rowcolor{gray!20}
\multicolumn{17}{c}{\bfseries \gellmo}
\\

\multirow{4}{*}{\gellmo}
 & 1 & \bestTurn{83.40}{0.60}{0.76} & 64.40 & 0.62 & 2.66 & \bestTurn{82.20}{0.60}{2.81} & \bestTurn{84.82}{0.59}{0.96} & \bestTurn{87.50}{0.59}{1.51} \\
 & 2 & 76.20 & 0.58 & 0.76 & \bestTurn{65.60}{0.59}{3.06} & 66.10 & 0.57 & \underline{3.33} & 70.16 & 0.58 & 1.03 & 73.96 & 0.57 & 1.69 \\
 & 3 & 74.40 & 0.58 & 0.75 & 62.80 & 0.58 & 3.12 & 63.56 & 0.56 & \textbf{3.37} & 67.02 & 0.57 & 1.04 & 68.75 & 0.56 & 1.41 \\
 & 4 & 74.40 & 0.58 & 0.74 & 61.00 & 0.58 & 3.26 & 62.71 & 0.56 & 3.21 & 66.49 & 0.58 & 0.98 & 70.83 & 0.56 & 1.38 \\

\addlinespace[3pt]

\multirow{4}{*}{\gellmoBaseRL}
 & 1 & \bestTurn{\underline{88.80}}{0.59}{0.83} & 69.80 & 0.62 & 3.01 & 88.14 & 0.59 & 2.87 & \bestTurn{90.05}{0.61}{0.99} & 92.71 & 0.59 & 1.44 \\
 & 2 & 82.40 & 0.58 & \textbf{0.88} & \bestTurn{\underline{81.60}}{0.59}{3.77} & \bestTurn{88.98}{0.58}{3.12} & 87.43 & 0.57 & \underline{1.11} & \bestTurn{\textbf{95.83}}{0.56}{\textbf{1.75}} \\
 & 3 & 80.00 & 0.58 & \underline{0.87} & 81.20 & 0.58 & \textbf{3.87} & 87.29 & 0.57 & 2.95 & 84.82 & 0.56 & \textbf{1.12} & \underline{94.79} & 0.55 & \underline{1.73} \\
 & 4 & 78.80 & 0.58 & \underline{0.87} & 79.60 & 0.58 & \underline{3.85} & 86.44 & 0.57 & 2.95 & 83.77 & 0.56 & 1.09 & 93.75 & 0.55 & 1.70 \\

\addlinespace[3pt]

\multirow{4}{*}{\gellmomfg}
 & 1 & \bestTurn{88.40}{0.60}{0.78} & 80.20 & 0.62 & 2.77 & 80.51 & 0.60 & 2.74 & 86.91 & 0.62 & 0.92 & 88.54 & 0.60 & 1.37 \\
 & 2 & 87.20 & 0.60 & 0.83 & \bestTurn{\textbf{83.60}}{0.61}{3.50} & 87.29 & 0.59 & 2.78 & \bestTurn{89.01}{0.60}{0.99} & 90.62 & 0.58 & 1.70 \\
 & 3 & 85.00 & 0.59 & 0.84 & 80.80 & 0.61 & 3.60 & 88.14 & 0.58 & 2.85 & 86.91 & 0.59 & 1.02 & \bestTurn{93.75}{0.58}{1.62} \\
 & 4 & 85.00 & 0.59 & 0.84 & 81.20 & 0.61 & 3.69 & \bestTurn{\underline{90.68}}{0.57}{2.87} & 85.34 & 0.59 & 1.04 & 91.67 & 0.58 & 1.42 \\

\bottomrule
\end{tabular*}
\begin{tablenotes}[flushleft]
\footnotesize
\item\hspace{-6pt}
The results with optimal $T$ for each model and task is highlighted in yellow.
Other metrics, notations, and formatting follow
Table \ref{tab:main_ind}.
\end{tablenotes}
\end{threeparttable}
\vspace{-8pt}
\end{table*}

\subsection{Implementation Details}
\label{app:implementation}
\paragraph{Training}
We use a group size of four for GRPO per prompt.
For groups with uninformative rewards, we apply dynamic resampling with up to three attempts.
We set the upper clipping parameter to $0.28$, the KL coefficient to $0.1$, and the maximum completion length to 96 tokens.
Training uses fused AdamW with $\beta_1=0.9$, $\beta_2=0.95$, $\epsilon=10^{-8}$, weight decay of $0.1$, and gradient clipping at $1.0$.
For the training data, we sample 100 source molecules for each tasks in \mumoinstruct, excluding the 5 OOD tasks.
Training GeLLM$^3$O-MFG took approximately 36 hours using two NVIDIA A100 GPUs.


For parameter-efficient training,
we use LoRA with rank 16, scaling parameter 32, and dropout 0.05.
We apply LoRA to the attention and Feed-Forward Network (FFN) projection layers in every Transformer block, 
while keeping all pretrained model parameters frozen.
Training of \gellmoBaseRL follows the same setup as \gellmomfg, but excludes generated context, feedback, and the reward bonus (Eq.~\ref{eq:reward_best}).
In terms of hyperparameters in reward function, 
$\kappa$ in Eq.~\ref{app:eq:property_improv} is set to $1.25$,
$\delta_{\mathrm{h}}$ in Eq.~\ref{eq:similarity_reward_details} is $0.6$ and $\delta$ in Eq.~\ref{reward_function} is $0.4$.

\paragraph{Inference}
As we discussed in Section~\ref{sec:experimental_setup:inference}, $N$ is set to $20/T$.
Candidate selection criterion uses the same $\delta$ and $\delta_{\mathrm{h}}$ values as training.

\subsection{Evaluation Metrics}
\label{app:evaluation_metrics}

\paragraph{Definition of \RI.}
Following \gellmo, $\RI$ is defined as follows,
\begin{equation}
    \RI = \frac{1}{|\propertyset|}\sum_{p \in \propertyset}\mathrm{RI}_p,
\end{equation}
where $\mathrm{RI}_p$ is the relative improvement on property $p$ and is defined as follows,
\begin{equation}
    \mathrm{RI}_p = d_p \frac{y_p(M)-y_p(S)}{y_p(S)}.
\end{equation}
$d_p$ is the direction coefficient and is set to $+1$ when property $p$ is optimized to increase and to $-1$ otherwise.

\subsection{Prompt and Feedback Templates}
\label{app:prompts}
All models receive the same initial prompt specifying the source molecule and target property objectives. The initial prompt follows the format:
%
%
\begin{quote}
\small
Your task is to modify the given molecule to adjust specific molecular properties while keeping structural changes as minimal as possible.
Your response should only contain a valid SMILES representation of the modified molecule enclosed with \texttt{<SMILES> </SMILES>} tag.\\
\texttt{\%\%\% Input: <SMILES>[source molecule]</SMILES>}\\
\texttt{\%\%\% Adjust: [property objectives]}
\end{quote}
%
%

In the MT setting,
each valid generation is followed by feedback on its structural similarity to the source molecule and improvements in each target property, as described in Section~\ref{sec:framework:problem_formulation}.
The feedback is structured as follows:
%
\begin{quote}
\small
\texttt{- Similarity to input molecule: [value] ([band])}\\
\texttt{- [property]: [source] -> [candidate]}\\
\texttt{([change], [status]; [direction])}
\end{quote}
Here, we categorize structural similarity into four bands: \texttt{very high} for similarity at least 0.6, \texttt{high} for $[0.5,0.6)$, \texttt{moderate} for $[0.4,0.5)$, and \texttt{low} below 0.4.
%

%

The feedback is also tailored based on whether a best molecule has been identified in previous turns.
If no best molecule exists, the model is instructed to improve properties that have not yet improved while preserving previously achieved improvements and structural similarity to $\source$.
Otherwise, the best molecule $C_t^*$ is directly provided in the prompt as a tagged SMILES string, guiding the model to further improve upon $C_t^*$.
%

For invalid generations, 
we provide corrective feedback specifying the reason for failure.
This allows the model to identify generation errors and refine its molecular generation strategy in subsequent turns.
%

\subsection{Additional Analysis of Experimental Results}
\label{app:results}
%
Tables~\ref{tab:all_turns_ind} and~\ref{tab:all_turns_ood} provide detailed performance results across different numbers of $T$, illustrating variations in the optimal $T$ across tasks and models.
Overall, \gellmomfg has a higher average optimal $T$ across the 10 tasks (2.70), compared with \gellmo (1.30) and \gellmoBaseRL (1.80).
This suggests that \framework trains \gellmo to effectively adapt its optimization strategies across turns, allowing improvements to accumulate over successive turns.
From $T=1$ to $T=4$, the average \sr of \gellmo decreases from $70.05\%$ to $61.64\%$, while \gellmoBaseRL improves slightly from $75.87\%$ to $75.94\%$. 
In contrast, \gellmomfg shows a more pronounced improvement from $76.24\%$ to $79.89\%$, demonstrating greater benefits from additional optimization turns.
These results suggest that \framework trains the model to better leverage multi-turn feedback for progressive molecular optimization, with a task-dependent optimal $T$.


\subsection{Artifact Sources and Licenses}
\label{app:artifact-licenses}

Table~\ref{tab:artifact-licenses} summarizes the sources,
licenses, and accessibility of the models, datasets, and
implementation used in our experiments. We report the
licenses specified by the respective artifact providers.
For artifacts without an explicitly stated license,
we indicate this accordingly.

\begin{table*}[t]
\centering
\caption{Licenses and sources of artifacts used in our study.}
\label{tab:artifact-licenses}

\begingroup
\footnotesize
\setlength{\tabcolsep}{5pt}
\renewcommand{\arraystretch}{1.3}

\begin{tabularx}{\textwidth}{
@{}p{0.22\textwidth} X
p{0.19\textwidth} p{0.13\textwidth}@{}
}
\toprule
\textbf{Artifact} &
\textbf{Source} &
\textbf{License Type} &
\textbf{Accessibility} \\
\midrule

MuMOInstruct &
\url{https://huggingface.co/datasets/NingLab/MuMOInstruct} &
CC BY 4.0 &
Dataset \\

Llama-3.1-8B-Instruct &
\url{https://huggingface.co/meta-llama/Llama-3.1-8B-Instruct} &
Llama 3.1 Community License &
Checkpoint \\

GeLLM$^3$O-P(6)$_{\mathrm{Llama}}$ &
\url{https://huggingface.co/NingLab/GeLLMO-P6-Llama} &
CC BY-NC 4.0 &
Checkpoint \\

Qwen3.5-9B &
\url{https://huggingface.co/Qwen/Qwen3.5-9B} &
Apache 2.0 &
Checkpoint \\

Ministral-3-8B-Instruct-2512-BF16 &
\url{https://huggingface.co/mistralai/Ministral-3-8B-Instruct-2512-BF16} &
Apache 2.0 &
Checkpoint \\

LlaSMol-Mistral-7B &
\url{https://huggingface.co/osunlp/LlaSMol-Mistral-7B} &
CC BY 4.0 &
LoRA Adapter \\

BioMedGPT-Mol &
\url{https://huggingface.co/PharMolix/BioMedGPT-Mol} &
Not specified &
Checkpoint \\

\bottomrule
\end{tabularx}

\endgroup
\end{table*}

\end{document}